%% file: main.tex
\documentclass{article}

\usepackage[preprint]{neurips_2025}
\usepackage{misc}
\usepackage{overpic}
\usepackage{hyperref}

\hypersetup{
  pdftitle={HyperDiffusionFields (HyDiF): Diffusion-Guided Hypernetworks for Learning Implicit Molecular Neural Fields},
  pdfauthor={Sudarshan Babu, Phillip Lo, Xiao Zhang, Aadi Srivastava, Ali Davariashtiyani, Jason Perera, Michael Maire, Aly A. Khan},
}

\title{HyperDiffusionFields (\hdf{}):\\Diffusion-Guided Hypernetworks for\\Learning Implicit Molecular Neural Fields}

\author{%
  Sudarshan Babu\thanks{Equal contribution.} \\
  CZ Biohub \\
  Chicago, IL \\
  \texttt{sudarshan.babu@czbiohub.org} \\
  \And
  Phillip Lo\footnotemark[1] \\
  CZ Biohub \\
  Chicago, IL \\
  \texttt{phillip.lo@czbiohub.org} \\
  \And
  Xiao Zhang\\
  University of Chicago\\
  Chicago, IL \\
  \texttt{zhang7@uchicago.edu} \\
  \And
  Aadi Srivastava \\
  Indian Institute of Technology Madras\\
  Chennai, TN \\
  \texttt{aadisrivastava.iitm@gmail.com} \\
  \And
  Ali Davariashtiyani\\
  University of Chicago\\
  Chicago, IL \\
  \texttt{davari@uchicago.edu} \\
  \And  
  Jason Perera \\
  CZ Biohub \\
  Chicago, IL \\
  \texttt{jason.perera@czbiohub.org} \\ 
  \And
  Michael Maire\\
  University of Chicago\\
  Chicago, IL \\
  \texttt{mmaire@uchicago.edu} \\
  \And
  Aly A. Khan \\
 University of Chicago\\ CZ Biohub Chicago\\
  Chicago, IL \\
  \texttt{aakhan@czbiohub.org} \\
}

\begin{document}
\maketitle

\begin{figure}[H]
   \centering
   \begin{overpic}[width=1.0\linewidth,grid=false]{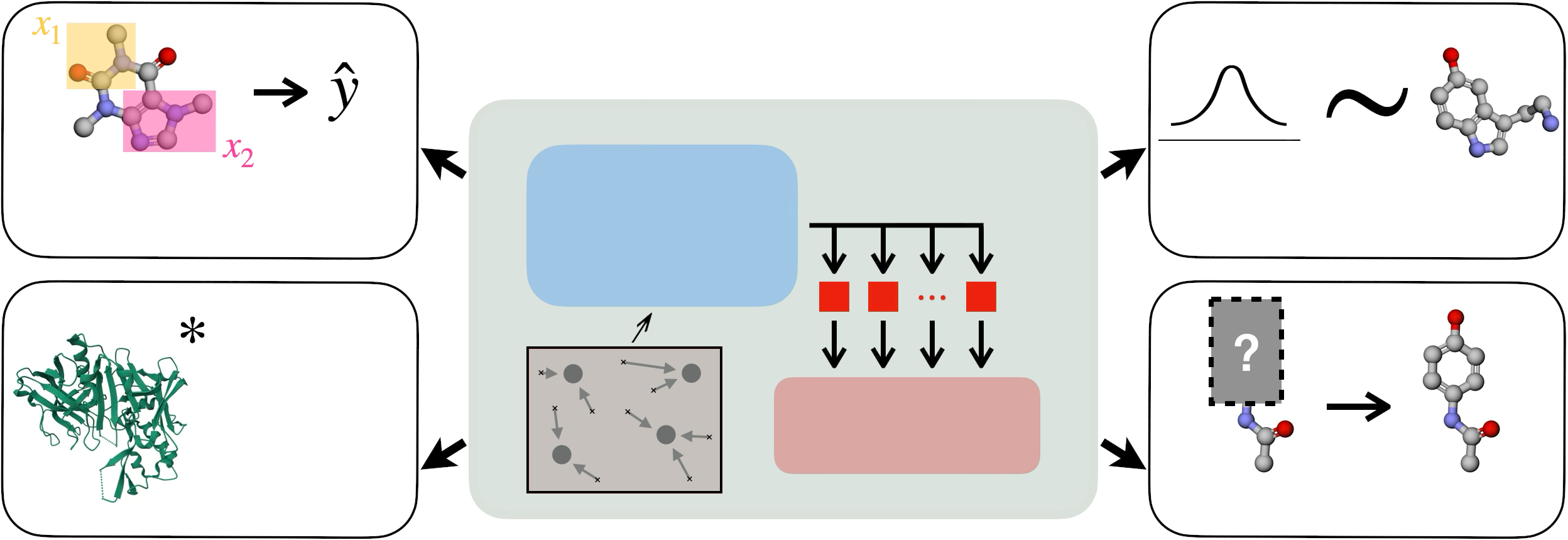}
      \put(5.5,21.5){\scriptsize{\textsf{localized features for}}}
      \put(6.25,19.5){\scriptsize{\textsf{property prediction}}}
      \put(14,9.5){\scriptsize{\textsf{scalability to}}}
      \put(16.5,7.5){\scriptsize{\textsf{larger}}}
      \put(13.75,5.5){\scriptsize{\textsf{biomolecules}}}
      \put(77,21.5){\scriptsize{\textsf{unconditional molecular}}}
      \put(82,19.5){\scriptsize{\textsf{generation}}}
      \put(78.5,2){\scriptsize{\textsf{molecular inpainting}}}
      \put(53.5,24){\scriptsize{\textbf{\textsf{Diffusion Model}}}}
      \put(36.25,19.5){\scriptsize{\textbf{\textsf{Hypernetwork}}}}
      \put(52.5,6.75){\scriptsize{\textbf{\textsf{Neural Field}}}}
      \put(46.5,29.5){\footnotesize{\textbf{\textsf{HyDiF}}}}
   \end{overpic}
   \caption[]{%
      HyperDiffusionFields (\hdf{}) is a framework for learning molecular conformers capable of a variety of tasks, including unconditional generation of molecules, inpainting, and representation learning for molecular property prediction. Furthermore, HyDiF scales to larger biomolecules.}
\end{figure}
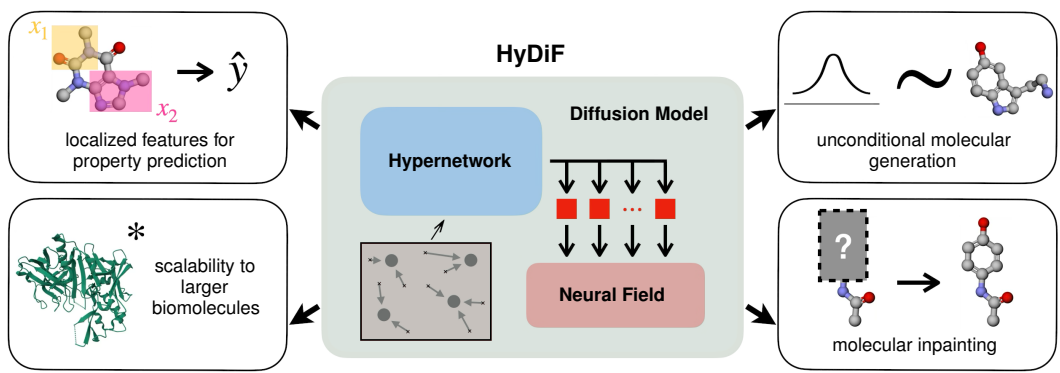

\begin{abstract}
    We introduce HyperDiffusionFields (\hdf{}), a framework modeling 3D molecular conformers as continuous fields rather than discrete atomic coordinates or graphs. At the core of our approach is the Molecular Directional Field (MDF), a vector field that maps any point in space to the direction of the nearest atom of a particular type. We represent MDFs using molecule-specific neural implicit fields, which we call Molecular Neural Fields (MNFs). To enable learning across molecules and facilitate generalization, we adopt an approach where a shared hypernetwork, conditioned on a molecule, generates the weights of the given molecule's MNF. To endow the model with generative capabilities, we train the hypernetwork as a denoising diffusion model, enabling sampling in the function space of molecular fields. Our design naturally extends to a masked diffusion mechanism to support structure-conditioned generation tasks, such as molecular inpainting, by selectively noising regions of the field. Beyond generation, the localized and continuous nature of MDFs enables spatially fine-grained feature extraction for molecular property prediction, something not easily achievable with graph or point cloud based methods. Furthermore, we demonstrate that our approach scales to larger biomolecules, illustrating a promising direction for field-based molecular modeling.\blfootnote{\textsuperscript{*}Protein structure from \cite{pdb-protein}.}
\end{abstract}

\input{introduction}
\input{related}

\input{method}

\input{results}

\input{discussion}

\medskip
\bibliographystyle{unsrtnat}
\bibliography{references}

\appendix
\include{supplement}

\end{document}

%% file: introduction.tex
\section{Introduction}
\label{sec:introduction}


Modeling complex, structured 3D data is a fundamental challenge in machine learning, with critical applications ranging from drug discovery and materials science to robotics and computer graphics \cite{bronstein2017geometric,qi2017pointnet,zhou2018voxelnet, zhao2021point, hanocka2019meshcnn,gnn2,merchant2023scaling}. Traditional approaches typically rely on discrete representations such as point clouds, meshes, or voxel grids. However, each of these come with limitations in resolution, expressiveness, or computational efficiency \cite{liu2025voxel, boulougouri2024molecular, wang2024prediction, wang2022point, shi2022deep}. To overcome this, we use implicit neural representations (INRs), which model continuous fields over 3D space using coordinate-conditioned neural networks, have shown remarkable success in capturing high-fidelity geometry and generalizing across spatial scales (e.g., SIREN \cite{siren}, NeRF \cite{mildenhall2021nerf}). By operating directly in continuous space, INRs are uniquely suited to capture fine-grained local geometric relationships crucial for modeling 3D objects \cite{kobayashi2022decomposing, lang2024iseg}.

Our work focuses on 3D molecular modeling, where precise geometric understanding is essential for tasks such as binding affinity prediction, docking and molecular design \cite{wang2024prediction,guan20233d}. Most existing approaches represent molecules as discrete objects—typically as graphs or point clouds—limiting the ability to capture continuous, fine-grained spatial relationships \cite{guan20233d,drotar2021structure}. We lift these discrete representations into continuous space via the Molecular Directional Field (MDF), which maps any point in space to the direction of the nearest atom of a given type. This dense, continuous representation encodes local geometry, provides high-resolution structural information, and offers robustness to perturbations by encoding correlated directional signals at nearby points \cite{kobayashi2022decomposing,lang2024iseg}.

\begin{figure}[t]
    \centering
    \includegraphics[width=\textwidth]{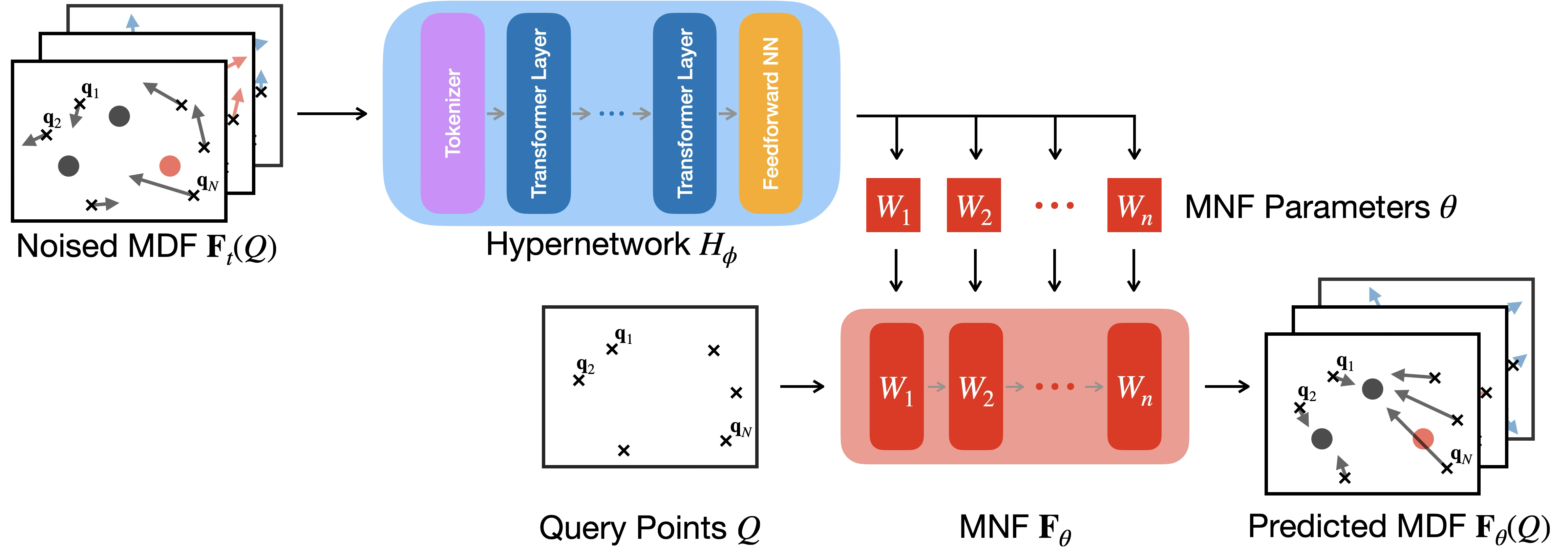}
    \caption{Overview of the \hdf{} architecture. A hypernetwork \(H_\phi\) takes a noised direction field \(\mathbf{F}_t(Q)\), query points \(Q\), and noising timestep \(t\), and produces the parameters \(\theta\) of an MNF \(\mathbf{F}_\theta\). The neural field is evaluated at the same query points to generate a denoised field, which is compared against the ground truth to compute the training loss. This enables \hdf{} to model molecules in a spatially localized manner.}
    \label{fig:block-diagram}
\end{figure}

To model the MDF, we propose \textbf{Hy}per\textbf{Di}ffusion\textbf{F}ields (\hdf{}, illustrated in Figure \ref{fig:block-diagram})---a generative framework for learning continuous molecular representations. At the core of our approach is the Molecular Neural Field (MNF), a coordinate-conditioned neural implicit that models the MDF. Since training a separate MNF for each molecule prevents learning shared structure across molecules, we adopt an approach similar to HyperFields \citep{babu2024hyperfields}, where a shared hypernetwork generates the parameters of per-molecule MNFs, enabling cross-molecular generalization. This adoption is further motivated by recent findings that hypernetworks often exhibit improved generalization in out-of-distribution (OOD) settings \cite{babu2024hyperfields, hypermaml, babu2025acquiringadaptingpriorsnovel, przewikezlikowski2024hypermaml}. To enable generative capabilities, we train the hypernetwork as a denoising diffusion model \cite{ddpm} over the space of MDFs, conditioning it to produce per-molecule MNFs. This allows explicit generation in the continuous function space of molecular fields, rather than discrete coordinate representations. 

Building on the model’s ability to capture fine-grained local geometry through the MDF, we extend our framework to support molecular inpainting, a critical task for \textit{in silico} drug discovery \cite{scaffold1,scaffold2,scaffold3,scaffold4}. This leverages the locality of our representation to enable structure-conditioned generation---a more practical and relevant capability than unconditional generation. The ability of \hdf{} to selectively regenerate parts of a molecule makes it well-suited for applications that require incorporating known structural constraints while completing or optimizing the rest.

In addition to generation, we utilize \hdf{} for feature extraction in downstream property prediction tasks. Generative models inherently learn rich, data-driven representations and recent work has demonstrated that such models excel at representation learning~\cite{zhang2025residual,zhang2023structural,zhang2024deciphering,yang2023diffusion,li2023mage}, enabling a unified framework that supports both generation and representation. Unlike traditional molecular encoders that produce a single global representation, our model naturally produces spatially localized features.
Finally, we demonstrate the scalability of our approach by applying \hdf{} to larger biomolecules such as proteins, which were previously intractable with existing methods. This extension underscores the flexibility of our framework in handling molecules of varying size and complexity. Overall, our work highlights a promising new direction in molecular modeling---one that shifts from discrete, rigid representations toward continuous, generative field-based models that unify representation learning and generation within a single framework.

    
    
    

%% file: related.tex
\section{Related Work}
In this section, we position our work relative to molecular representations, molecular generative and inpainting models, implicit neural representations, and hypernetworks.

\subsection{Molecular Representations}
Traditional approaches often represent molecules as strings (e.g., SMILES \cite{smiles1,smiles2,smiles3}, SELFIES \cite{selfies1,selfies2}, InChI \cite{inchi}) or graphs that capture atom connectivity but treat 3D structure as auxiliary. To explicitly incorporate spatial information, recent models use 3D point clouds \cite{edm,geoldm,midi} or voxel grids \cite{voxmol}. However, point clouds are sparse and permutation-sensitive, while voxel grids are resolution-limited and memory-intensive.

To address these limitations, we introduce the Molecular Directional Field (MDF)—a continuous vector field that maps 3D coordinates to atom-directed vectors, enabling dense, localized encoding of molecular geometry. This representation supports spatial querying, interpretability, and serves as a robust inductive bias for generative modeling.

\subsection{Models for 3D Molecular Generation}

Existing methods employ variational autoencoders (VAEs) \cite{vae1,vae2,vae3}, normalizing flows \cite{nf1,nf2,nf3}, or generative adversarial networks (GANs) \cite{gan1,gan2,gan3} over strings or molecular graphs. These models often lack fine-grained control over 3D geometry, a critical limitation for tasks like conformer generation or structure-based drug design \cite{sbdd1,sbdd2,confgen1,confgen2,confgen3}.

Denoising diffusion models have recently advanced the state-of-the-art in 3D molecular generation, with methods such as EDM \cite{edm}, GeoLDM \cite{geoldm}, and MiDi \cite{midi}, applying noise to conformers and denoising in coordinate or latent spaces. These models often rely on equivariant architectures, but they still operate on sparse atomic representations. VoxMol \cite{voxmol} instead uses voxelized occupancy grids, providing a denser spatial representation, though at the cost of resolution and memory efficiency.

FuncMol \cite{funcmol} models 3D molecules using neural occupancy fields and performs diffusion in a learned latent space, capturing global molecular geometry while operating on compressed representations. Their approach requires a two-stage training process---first learning the field representation, then training a separate diffusion model in latent space. 
Another related model MCF (Molecular Conformer Fields \cite{mcf}, \cite{wang2024swallowing}) approaches conformer generation by learning a distribution over functions that map molecular graph elements to 3D coordinates using diffusion.

In contrast to these approaches, \hdf{} performs diffusion directly in the function space of Molecular Directional Fields (MDFs)---dense, spatially continuous vector fields that map arbitrary 3D points to atom-directed vectors. Unlike latent-based methods, our framework is trained end-to-end and supports localized generation via masked denoising. While MCF models a mapping from molecular graphs to conformers and is limited to graph-to-structure generation, \hdf{} models a function from 3D space to directional vectors, enabling both generation from noise and spatially localized structure-conditioned editing.

\subsection{Molecular Inpainting and Scaffold-Constrained Generation}
Scaffold-based methods like ScaffoldGVAE \cite{scaffoldgvae}, Sc2Mol \cite{sc2mol}, and MoLeR \cite{moler} operate on SMILES or molecular graphs, generating molecules by conditioning on a core scaffold and elaborating it via learned motifs or side-chain transformations. Others such as DiffSBDD \cite{sbdd1} perform 3D inpainting in the context of protein-ligand binding. 

In contrast, \hdf{} performs localized editing directly in continuous 3D space via masked denoising on Molecular Directional Fields. Unlike prior work, our edits are not constrained to predefined scaffolds or external protein contexts; instead, we enable user-specified, spatially targeted edits by masking arbitrary regions in the molecular field, offering a flexible, geometry-aware approach to molecular structure manipulation.

\subsection{Hypernetworks}
Hypernetworks \cite{ha2016hypernetworks} are neural networks that generate the weights of another network, enabling dynamic and flexible parameterization. They have proven particularly effective for generating implicit neural representations (INRs), where a shared hypernetwork maps latent codes to the weights of coordinate-based MLPs that model continuous signals \cite{babu2024hyperfields, chen2022transformers}. Additionally, works highlight that hypernetworks facilitate out-of-distribution generalization by decoupling representation learning from signal generation \cite{hypermaml, przewikezlikowski2024hypermaml}. Hence, we use hypernetworks in \hdf{} to model a distribution over implicit molecular fields. This enables parameter sharing across molecules, supports generation and inference over unseen structures, and allows us to train on large libraries of molecule-specific neural implicits in a unified framework.

\subsection{Generative Models as Feature Extractors}

Recent work in computer vision has shown that diffusion models can serve as effective feature extractors, with intermediate activations capturing rich semantic information necessary for generation \cite{diff-beat-gan, diffusion-hyperfeatures, zhang2023structural}. Motivated by this, we explore the use of intermediate representations from \hdf{} for downstream molecular tasks. This dual role---both generative and descriptive---makes \hdf{} the first model in molecular conformer modeling to unify generation and feature extraction within a single framework.

%% file: method.tex
\section{Method}
In this section, we describe the \hdf{} framework. Each molecule is represented as a dense field that maps every point in 3D space to a vector pointing toward the nearest atom, yielding a locally structured and geometrically aware representation. Our model learns to generate such fields by reversing a diffusion process using a transformer-based hypernetwork that predicts the parameters of a neural field conditioned on noisy observations. This formulation supports both unconditional generation and molecular inpainting, and enables local feature extraction from molecular structure. \hdf{} brings together diffusion models for flexible generative modeling, hypernetworks for task-specific adaptivity, and implicit neural representations for capturing fine-grained 3D molecular structure.

\subsection{Molecular Direction Fields (MDFs)}
We represent 3D molecular conformers using \textit{Molecular Direction Fields}—vector fields over continuous space that capture local geometric structure. At every point in space, the field points toward the nearest atom in the molecule, offering a dense representation of molecular shape.

To build intuition for MDFs, consider constructing a direction field for a toy molecule consisting of atoms of a single type---e.g., carbon. Let \(\mathcal{A} = \{\mathbf{a}_1, \dots, \mathbf{a}_n\}\) denote the set of atomic coordinates. We define the vector field \(\mathbf{F} : \mathbb{R}^3 \to \mathbb{R}^3\) where for any query point \(\mathbf{q} \in \mathbb{R}^3\), the output \(\mathbf{F}(\mathbf{q})\) is the vector pointing from \(\mathbf{q}\) to the nearest atom in \(\mathcal{A}\):
\begin{equation}
\mathbf{F}(\mathbf{q}) = \mathbf{a}_{\textrm{nearest}} - \mathbf{q},\;\textrm{where}\; \mathbf{a}_{\textrm{nearest}} = \arg\min_{\mathbf{a} \in \mathcal{A}} \|\mathbf{q} - \mathbf{a}_i\|_2.
\label{eqn:direction-field}
\end{equation}

To extend to molecules with multiple atom types, we construct a separate direction field \(\mathbf{F}^{(k)}\) for each atom type \(k\) over the subset \(\mathcal{A}_k\subset \mathcal{A}\) which contains only atoms of type \(k\). The resulting representation is a multi-channel vector field \(\mathbf{F}: \mathbb{R}^3 \to \mathbb{R}^{K \times 3}\), where \(K\) is the number of distinct atom types in the dataset. Each channel \(\mathbf{F}^{(k)}(\mathbf{q})\) is the vector from \(\mathbf{q}\) to the nearest atom of type \(k\).

To extend to molecules with multiple atom types, we construct a separate direction field \(\mathbf{F}^{(k)}\) for each atom type \(k\) over the subset \(\mathcal{A}_k\subset \mathcal{A}\) which contains only atoms of type \(k\). The resulting representation is a multi-channel vector field \(\mathbf{F}: \mathbb{R}^3 \to \mathbb{R}^{K \times 3}\), where \(K\) is the number of distinct atom types in the dataset. Each channel \(\mathbf{F}^{(k)}(\mathbf{q})\) is the vector from \(\mathbf{q}\) to the nearest atom of type \(k\).

\begin{SCfigure}[1]
    \includegraphics[width=0.6\textwidth]{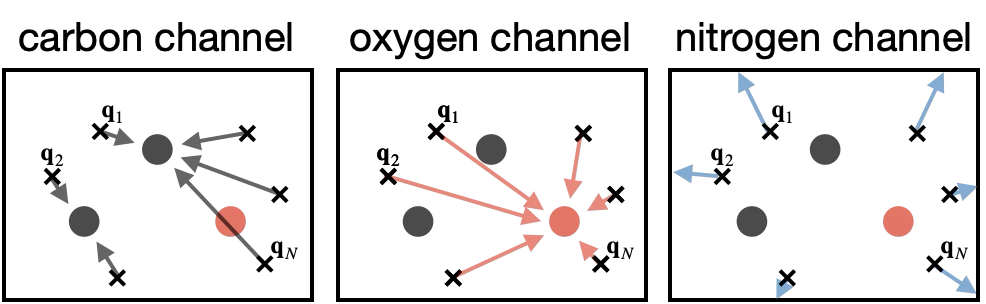}
    \caption{Direction field channels for a molecule with two carbons, one oxygen, and no nitrogens. Each query point is mapped to the nearest atom of the corresponding type. Observe the nitrogen channel points outwards.}
    \label{fig:mdf-illustration}
\end{SCfigure}

Each molecule in our dataset is thus represented by a \(K\)-channel direction field, one channel for each possible atom type. If a particular type \(k\) is not present in a given molecule, we populate that channel with outward-pointing vectors that radiate from the molecule’s center of mass toward a bounding sphere. This ensures that the representation remains dense and consistent across all molecules. This construction is illustrated in Figure~\ref{fig:mdf-illustration}.

While MDFs provide a rich representation for molecules, for downstream applications we are still interested in human-parsable graph or string representations; we describe how we recover molecular graphs from MDFs in \S\ref{sec:reconstruction} in the supplementary material.

\subsection{Molecular Neural Fields (MNFs)}

We represent Molecular Direction Fields using \textit{Molecular Neural Fields} (MNFs)—coordinate-based neural networks trained to approximate the ground truth MDF of a molecule. Each MNF is a function \(\mathbf{F}_{\theta} : \mathbb{R}^3 \to \mathbb{R}^{K \times 3}\) that maps a query point to a multi-channel output, where the \(k\)th channel predicts the direction to the nearest atom of type \(k\). The network is trained to match $\mathbf{F}_{\theta}$ to $\mathbf{F}$ at sampled points and is parameterized as an MLP with sinusoidal activations known as a SIREN \cite{siren}, which is well-suited for modeling high-frequency geometric signals.

\subsection{Forward Diffusion Process}
\label{sec:forward-diffusion}

To model a distribution over Molecular Direction Fields, we adopt the denoising diffusion probabilistic model (DDPM) framework \cite{ddpm}, applied to vector fields. The forward diffusion process defines a Markov chain that gradually transforms a clean field into Gaussian noise over \(T\) discrete steps.

Given a ground-truth direction field \(\mathbf{F}\), evaluated at a set of spatial query points \(Q = \{\mathbf{q}_j\}_{j=1}^N\), we construct the noised field at timestep \(t\) as:
\begin{equation}
\mathbf{F}_t(Q) = \sqrt{\bar{\alpha}_t} \cdot \mathbf{F}(Q) + \sqrt{1 - \bar{\alpha}_t} \cdot \boldsymbol{\epsilon}, \quad \boldsymbol{\epsilon} \sim \mathcal{N}(0, I),
\label{eqn:fwd-noising}
\end{equation}
where \(\bar{\alpha}_t = \prod_{i=1}^{t}(1 - \beta_i)\), and \(\{\beta_i\}_{i=1}^T\) is a fixed noise schedule. We adopt the cosine schedule proposed in \cite{cos-schedule}, which improves stability over the linear schedule in high-resolution spatial domains.

\begin{SCfigure}[1]
    \includegraphics[width=0.55\textwidth]{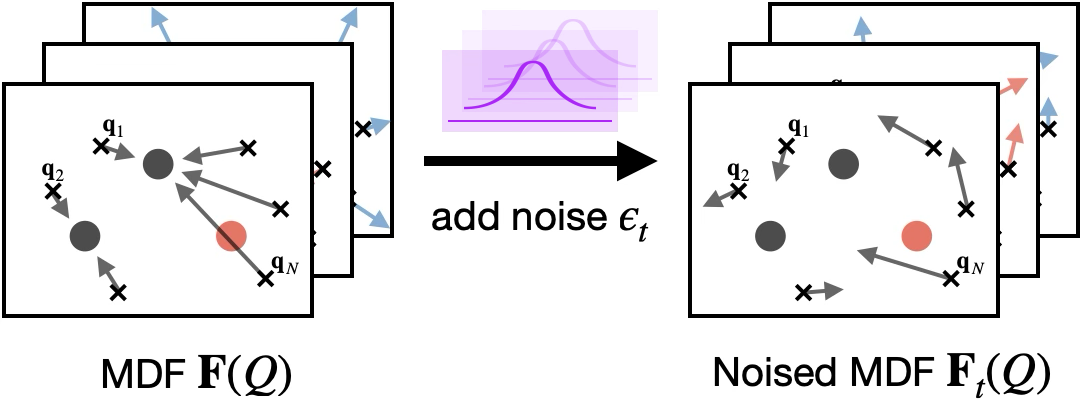}
    \caption{The forward noising process for an MDF, where i.i.d. Gaussian noise is added to each component of each vector at each query point. Here, \(\boldsymbol{\epsilon}_t\) is short for \(\sqrt{1 - \bar{\alpha}_t} \cdot \boldsymbol{\epsilon}\).\vspace{3.5em}}
    \label{fig:mdf-noising}
\end{SCfigure}

At timestep \(t = 0\), the noised field reduces to the clean field, while at the final step \(t = T\), the noised field \(\mathbf{F}_T(Q)\) becomes almost indistinguishable from isotropic Gaussian noise. This progression forms a trajectory through field space, allowing the model to learn to denoise under varying signal-to-noise ratios. The noise is applied independently to each vector component at each spatial location and for each channel in the field. Each vector \(\mathbf{F}_t(\mathbf{q})\) is corrupted by additive Gaussian noise scaled according to the diffusion schedule (see Figure \ref{fig:mdf-noising}).

\subsection{Architecture}
\label{sec:methods-architecture}
The core idea of \hdf{} (see Figure \ref{fig:block-diagram}) is to use a hypernetwork to generate the parameters of a MNF that denoises a noised molecular direction field. The hypernetwork, denoted \( H_\phi \), takes three inputs: the noised direction field at timestep \( t \), denoted \( \mathbf{F}_t(Q) \); the corresponding 3D query points \( Q = \{\mathbf{q}_j\}_{j=1}^N \); and the diffusion timestep \( t \) itself. The output is a set of parameters \( \theta \), which define a coordinate-based neural network \(\mathbf{F}_\theta \) that serves as the denoised MNF.  Formally, we have 
\begin{equation}
    \theta = H_\phi(\mathbf{F}_t(Q), Q, t).
    \label{eqn:hypernetwork}
\end{equation}
We include further details about the \hdf{} architecture in \S\ref{sec:implementation-details} of the supplementary material.

\subsection{Reverse Process and Training}
\label{sec:methods-training}
Once the parameters \( \theta \) are generated by the hypernetwork, we use the resulting MNF \( \mathbf{F}_\theta \) to reverse the forward noising process. Specifically, we evaluate \( \mathbf{F}_\theta\) at the same query points used in the forward process to predict the clean direction field. These predictions are then compared to the ground-truth direction vectors in the following training objective:
\begin{equation}
\min_\phi \,\mathcal{L}_{\textrm{vec}}, \;\textrm{where}\; \mathcal{L}_{\textrm{vec}}= \frac{1}{|Q|} \sum_{\mathbf{q} \in Q} \sum_{k=1}^K \left\| \mathbf{F}_\theta^{(k)}(\mathbf{q}) - \mathbf{F}^{(k)}(\mathbf{q}) \right\|_1 \;\textrm{and}\;\theta = H_\phi(\mathbf{F}_t(Q), Q, t).
\end{equation}
In practice, we find this vector-based objective to be unstable. Molecular direction fields often contain high-frequency discontinuities, particularly where the identity of the nearest atom can change abruptly across neighboring spatial points. These discontinuities make the direction field difficult to fit with compact MLPs, especially when training with noisy 
inputs. To address this, we instead train the MNF to predict the scalar \textit{distance field}, defined as the Euclidean distance from a point \(\mathbf{q}\) to the nearest atom of each type. Define
\begin{equation}
f^{(k)}(\mathbf{q}) = \|\mathbf{a}^{(k)}_{\textrm{nearest}} - \mathbf{q}\|,\; \textrm{where}\; \mathbf{a}^{(k)}_{\textrm{nearest}} = \arg\min_{\mathbf{a} \in \mathcal{A}_k} \|\mathbf{q} - \mathbf{a}\|_2.
\end{equation}
This scalar function is smoother and easier for neural networks to approximate. We correspondingly define the MNF as a function \(f_\theta: \mathbb{R}^3 \to \mathbb{R}^K\), where each output channel predicts the distance to the nearest atom of type \(k\). The associated training objective is then 
\begin{equation}
\min_\phi \,\mathcal{L}_{\textrm{dist}}, \;\textrm{where}\; \mathcal{L}_{\textrm{dist}} = \frac{1}{|Q|} \sum_{\mathbf{q} \in Q} \sum_{k=1}^K \left| f^{(k)}_\theta(\mathbf{q}) - f^{(k)}(\mathbf{q}) \right|\;\textrm{and}\;\theta = H_\phi(\mathbf{F}_t(Q), Q, t).
\end{equation}
In this setup, each MNF predicts one scalar distance per atom type for every spatial location. Through experimentation, we observe that the best performance is achieved when the hypernetwork is \textit{conditioned on} a noised direction field, but trained to \textit{output} a distance field. This hybrid setup leverages the high-frequency information present in the direction field for conditioning, while taking advantage of the smoother, lower-frequency distance field as the target for regression. We provide ablation studies supporting this choice in \S\ref{sec:ablation} in the supplementary material. Full algorithms for forward noising and generation are provided in the supplementary material \S\ref{sec:algos}. 


\subsection{Masked Diffusion Procedure for Molecular Inpainting}

\label{sec:method-masked-diff}
 The geometric nature of the \hdf{} architecture supports a molecular inpainting task: given a partially masked molecule, the model completes the missing regions, enabling selective modification of portions of a molecule. This is a crucial step in the drug discovery pipeline commonly referred to as \textit{lead optimization} \cite{leadoptim1,leadoptim2,leadoptim3,leadoptim4}. We discuss the details of our masked diffusion procedure more in \S\ref{sec:supp-masked-diff}.


%% file: results.tex
\section{Results}
\begin{figure}
    \centering
    \includegraphics[width=0.6\linewidth]{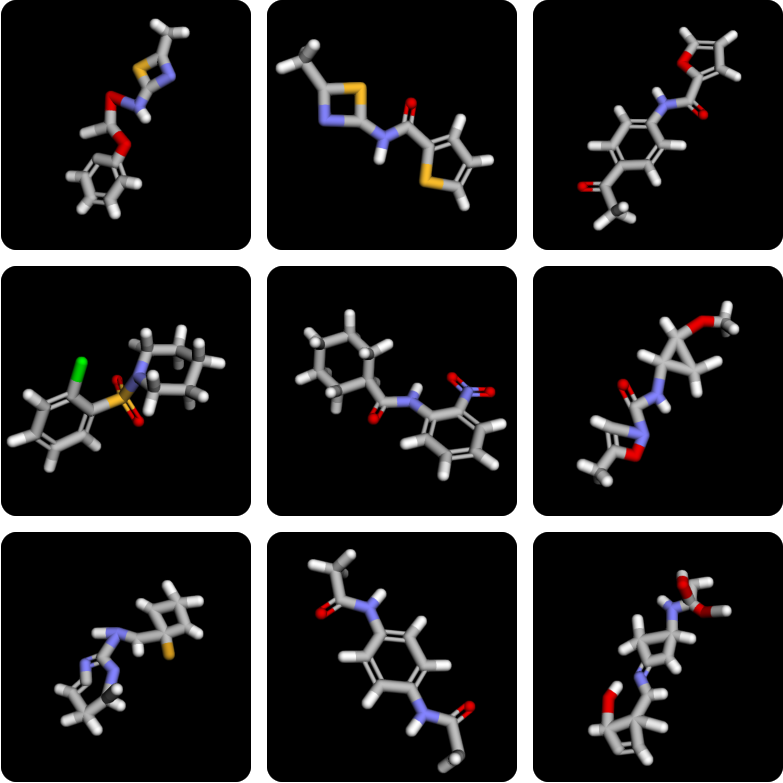}
    \caption{Nine curated samples from unconditional generation. \hdf{} generates chemically plausible and structurally diverse molecules when sampling unconditionally.}
    \label{fig:unconditional-gallery}
\end{figure}

We pretrain \hdf{} on the gold standard GEOM (Geometric Ensemble of Molecules \cite{geom}) dataset—a collection of high‑quality conformers computed \textit{in silico} using semi‑empirical tight‑binding density functional theory. We use the same GEOM‑drugs dataset and splits as \cite{funcmol}, discarding molecules containing elements other than C, H, O, N, F, S, Cl, and Br; this yields a training set of 1.1M conformers across 242K species. We show unconditional generation results after pretraining in Figure \ref{fig:unconditional-gallery}, where nine curated samples demonstrate that \hdf{} generates diverse and realistic conformers when sampling from pure noise. In \S\ref{sec:protein-scaling} of the supplementary material, we describe and show preliminary results of fitting \hdf{} to proteins.

\subsection{Molecular Inpainting}

In the drug discovery pipeline, a candidate drug for a particular task is known as a lead. Medicinal chemists are then interested in finding a large number of structural analogues to this lead, i.e., a variety of molecules with similar structure, in hopes of having a large number of candidates with similar interaction with a biological target, but perhaps more favorable absorption, etc. For this reason, the conditional generation task (molecular inpainting) is much more relevant in real-world applications, as opposed to the unconditional task that most other works evaluate their models on. 

\begin{figure}
    \centering
    \includegraphics[width=0.6\textwidth]{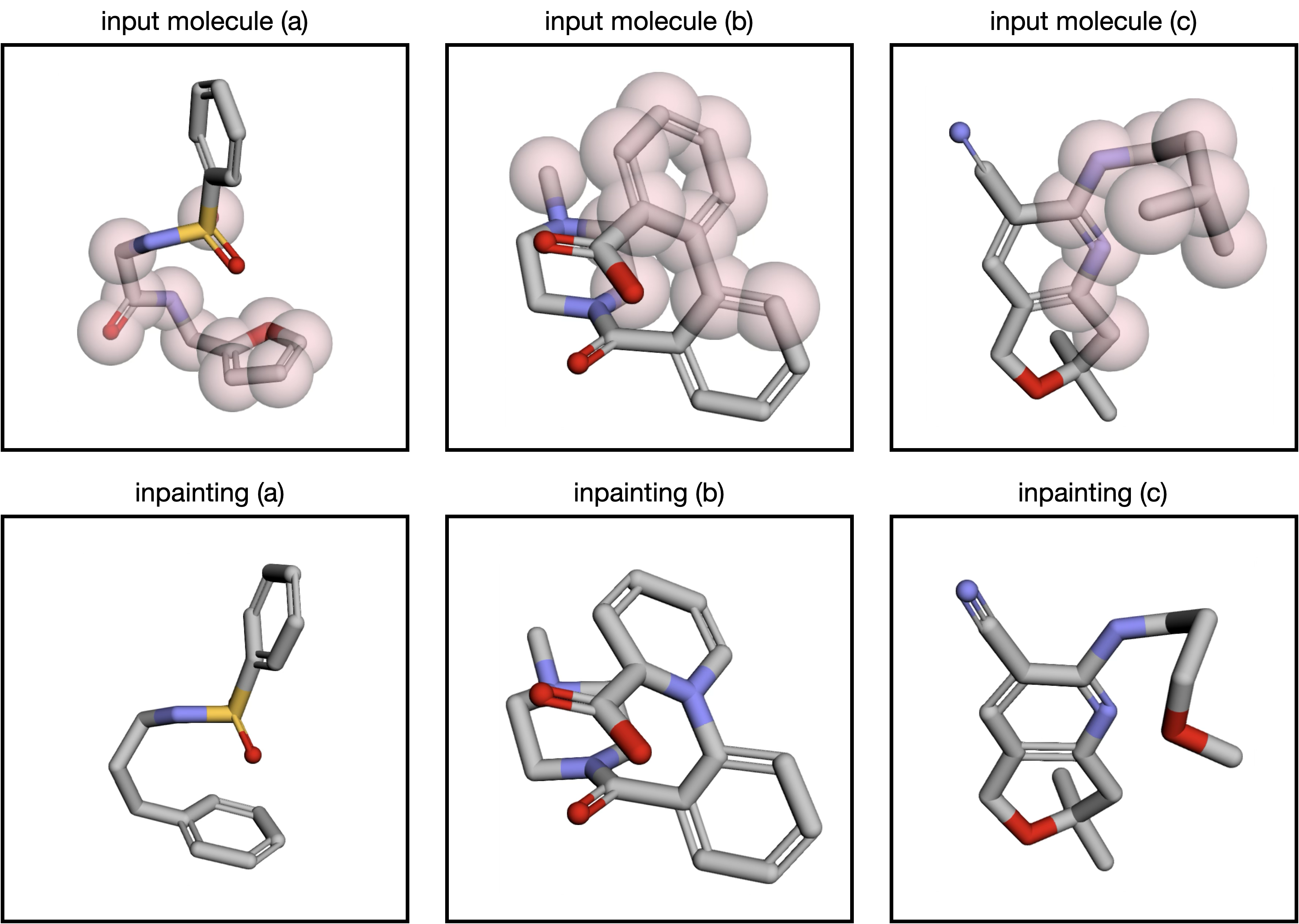}
    \caption{Three molecules from the validation set and their corresponding inpaintings. Yellow denotes the region of the molecule we selected for inpainting. The above examples demonstrate the ability of \hdf{} to inpaint effectively. We attribute this to \hdf{}'s ability to locally model conformers via MDFs. (a)  replacement of the five-member ring with a six-member ring, and the replacement of a nitrogen (blue) with carbon (gray). (b) replacement of a carbon in a benzene ring with a nitrogen. (c) replacement of a branched chain with a non-branched chain. }
    \label{fig:inpainting-carousel}
\end{figure}

We train \hdf{} on the molecular inpainting task similarly to the pretraining method in \S\ref{sec:methods-training} but with the masked diffusion procedure in \S\ref{sec:supp-masked-diff}. In Figure \ref{fig:inpainting-carousel}, we exhibit a selection of partially masked conformers from the validation set and the corresponding inpainted molecules. In each case, \hdf{} is able to successfully perform local edits on a molecule while maintaining chemical validity.

To quantify the quality of inpainting, we compute three metrics on generated conformers: \textit{atom stability}, \textit{molecule stability}, and \textit{validity}. Atom stability measures the percentage of atoms with correct valency; molecule stability measures the percentage of molecules where all atoms have correct valency; validity reports the fraction of molecules that pass RDKit's sanitization check. Table~\ref{tab:inpainting-metrics-main} compares \hdf{} to the EDM baseline on these metrics. 

The inpainting task highlights the utility of our novel field-based representation of molecules and our accompanying hypernetwork architecture, which performs diffusion directly in molecular space rather than latent space; this is a departure from most other methods. This is why we are only able to compare to EDM (which performs diffusion in atomic coordinates space); most other models such as FuncMol and GeoLDM perform diffusion in latent space. Latent diffusion models cannot be straightforwardly modified for the conditional generation task. On the other hand, for EDM and our method, the diffusion in molecular space makes it straightforward to perform diffusion on a partially masked molecule.

\begin{table}
\centering
\caption{Molecular inpainting metrics on GEOM (higher is better). Metrics for EDM and \hdf{} are computed over 100 generated conformers. \hdf{}'s high-resolution molecular representation is able to inpaint more chemically reasonable molecules than EDM's discrete representation.}
\captionsetup{width=0.8\linewidth}  
\begin{tabular}{@{}lccc@{}}
\toprule
       & Atom Stable \% & Mol. Stable \% & Validity \% \\
\midrule
data &       100.0\%         &   100.0\%     &    96.9\% \\
\midrule
EDM    &        93.1\%        &      13\%          &    85\%        \\
\hdf{}  &   \textbf{95.1}\%             &   \textbf{36}\%    &   \textbf{93}\%      \\
\bottomrule
\end{tabular}
\label{tab:inpainting-metrics-main}
\end{table}

\subsection{Molecular Property Prediction}
\label{sec:results-property-prediction}
\textbf{Datasets.} To evaluate how well our pretrained model generalizes to downstream property prediction tasks, we use a subset of benchmarks from MoleculeNet \cite{moleculenet}, which span regression and binary classification tasks (see \S \ref{sec:property-prediction-tasks} in the supplementary material for details).

\textbf{Baselines.} We compare our method against a large-scale sequence-based foundational model and two structure-based diffusion models trained on the same dataset as ours. ChemBERTa-2 \cite{chemberta} is a RoBERTa-style \cite{roberta} transformer pretrained on 77M SMILES strings using masked language modeling. As a foundational model trained on orders of magnitude more data than ours, it serves as a strong baseline for molecular property prediction. In contrast, EDM \cite{edm} and GeoLDM \cite{geoldm} are pretrained diffusion models trained on GEOM. Both operate directly on 3D molecular structures using graph neural networks: EDM employs a SE(3)-GNN, while GeoLDM uses SE(3)-GNN with latent diffusion. To repurpose EDM and GeoLDM as feature extractors, we extract a global molecular embedding by averaging per-atom feature vectors from intermediate layers of the GNN.

\textbf{{\hdf{}}.} \hdf{} is pretrained on the GEOM dataset using the procedure described earlier in \S\ref{sec:methods-training}. For a given molecule, we generate its MDF and sample a dense set of query points uniformly throughout the molecular volume. The MDF is passed through a pretrained \hdf{}, from which we extract intermediate activations as features. This allows us to obtain localized representations at high spatial resolution. We average these local features to produce a single global embedding for the molecule.

\textbf{{\hdf{}} + Chemberta-2.} In addition to using \hdf{} on its own, we also evaluate a fusion model that concatenates the global embedding from \hdf{} with ChemBERTa-2's representation. This combined feature vector is then passed to the same MLP head for downstream property prediction. This allows us to test the complementarity of geometry-aware and sequence-based representations.

\textbf{Comparison.} In Table \ref{tab:property-prediction-results}, across all tasks, \hdf{} compares favorably to both ChemBERTa-2 and the structure-based diffusion baselines. This is despite the fact that ChemBERTa-2 is trained on 77M unique species, while \hdf{} is trained on 242K unique species. We attribute this performance to a fundamental difference in how our model represents molecules: rather than producing a single global embedding directly, \hdf{} aggregates localized features learned from spatially resolved query points. This allows the model to capture fine-grained geometric information that is otherwise lost in global or sequence-based representations. 



\begin{table}
    \centering
    \caption{Comparison on MoleculeNet tasks. \hdf{} alone compares favorably with baselines which learn global representations, highlighting the efficacy of locally learned representations.}
    \begin{tabular}{@{}l |cc|ccc@{}}
    \toprule
    & \multicolumn{2}{c|}{\textbf{Classification (AUROC $\uparrow$)}} & \multicolumn{3}{c}{\textbf{Regression (RMSE $\downarrow$)}} \\
                  & BBBP & BACE  & Lipo & ESOL & FreeSolv \\
    \midrule
    ChemBERTa-2                        & 0.70 &   0.79        & 0.81 & 0.98 & 2.12 \\
    \midrule
    EDM                                & 0.67 &  0.78      & 0.94 & 1.03 & 2.25   \\
    GeoLDM                             & 0.68 &  0.78       & 0.97 & 1.26 & 2.60       \\
    \midrule
    \hdf{}               & 0.71 &  0.79       & 0.83 & 0.90 & 1.82    \\
    \hdf{} + ChemBERTa-2 & \textbf{0.73} &  \textbf{0.80}       & \textbf{0.78} & \textbf{0.89} & \textbf{1.75}    \\
    \bottomrule
    \end{tabular}
    \label{tab:property-prediction-results}
\end{table}

%% file: discussion.tex
\section{Conclusion}

This work introduces \hdf{}, a generative framework that models 3D molecular structure via direct diffusion on continuous vector fields and neural function spaces. The continuous nature of our MDF representation allows for \hdf{} to learn rich localized representations, enabling highly competitive feature generation for  molecular property prediction as well as the ability to make localized edits to molecules via molecular inpainting. Moreover, the coordinate- and grid-free nature of our representation allows \hdf{} to scale to larger biomolecules. We believe the contributions of \hdf{} represent a promising direction for future work in geometric deep learning, one that bridges the gap between high-fidelity representation and flexible, spatially aware generation.

%% file: supplement.tex

\section{Pretraining Results}
\label{sec:pretraining-results}
\begin{figure}
    \centering
    \includegraphics[width=0.8\linewidth]{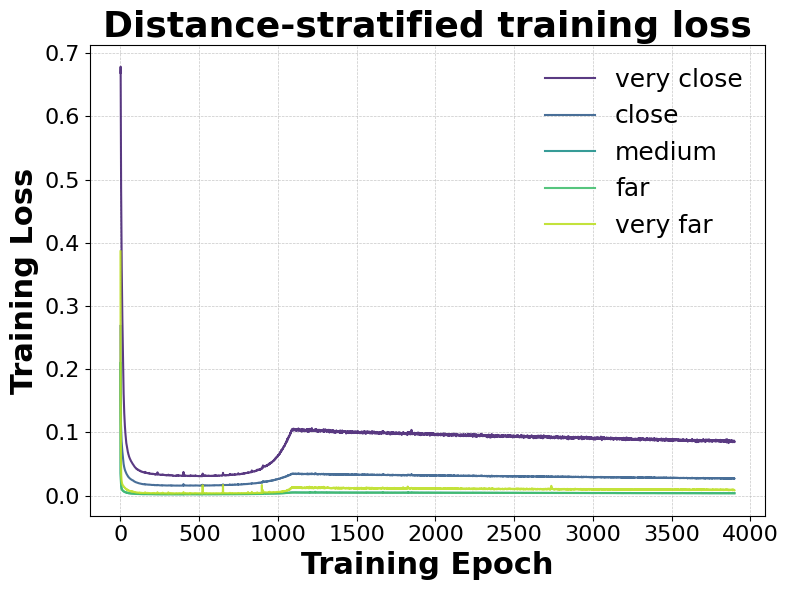}
    \caption{Training curves stratified by the distance from the query point to the nearest atom. Loss is tracked separately for five distance bins: \emph{very close}, \emph{close}, \emph{medium}, \emph{far}, and \emph{very far}. The \emph{very close} regime is the most challenging due to the high-frequency variation near the molecular surface. Loss in each bin initially decreases, then increases due to the progressive diffusion noise schedule.}
    \label{fig:pretraining-curves}
\end{figure}

\begin{figure}
\centering
\includegraphics[width=0.8\linewidth]{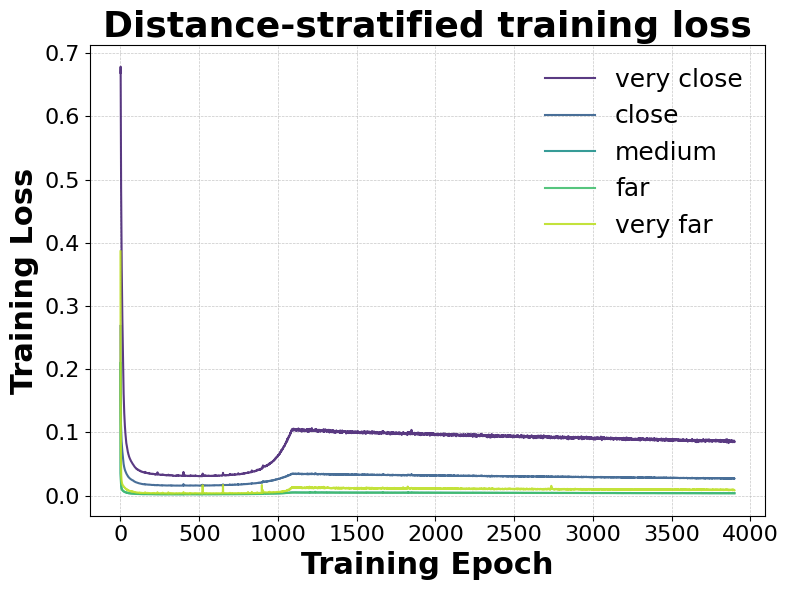}
\caption{Training curves stratified by the distance from the query point to the nearest atom. Loss is tracked separately for five distance bins: \textit{very close}, \textit{close}, \textit{medium}, \textit{far}, and \textit{very far}. The \textit{very close} regime is the most challenging due to the high-frequency variation near the molecular surface. Loss in each bin initially decreases, then increases due to the progressive diffusion noise schedule.}
\label{fig:unconditional-curves}
\end{figure}

In Figure~\ref{fig:unconditional-curves}, we show distance-stratified training curves that break down the training loss based on the proximity of each query point to its nearest atom. Specifically, we divide query points into five distance bins: \textit{very close}, \textit{close}, \textit{medium}, \textit{far}, and \textit{very far}; the percentile bins are given by $[0\%-2\%], [2\%-33\%], [33\%-66\%], [66\%-98\%], [98\%-100\%]$.  For each bin, we track the training loss separately over time.

These curves reveal that the \textit{very close} category is the most challenging to learn. This regime contains query points nearest the atoms of the molecule, where high-frequency geometric variation is most pronounced. In such regions, the direction to the atom can change abruptly across small distances, introducing sharp changes in the direction field. Accurately modeling these points requires the network to capture fine-grained structural detail. As such, improvements in the \textit{very close} bin are strong indicators that the model is learning the geometry of the molecule with high fidelity.

Observe that the loss in all bins initially decreases but then rises again during training before plateauing. This behavior is due to the training curriculum we employ, in which the diffusion noise level is gradually increased across epochs. Early in training, the model sees easier (low-noise) examples, leading to rapid improvements. As training progresses, higher levels of noise are introduced, making the task more difficult and increasing the overall loss. We describe this curriculum in more detail in Section~\ref{sec:implementation-details}.

\section{Implementation Details}
\label{sec:implementation-details}

\textbf{Hypernetwork Implementation Details.}  
As described in Equation~\ref{eqn:hypernetwork} and Section~\ref{sec:methods-architecture}, the hypernetwork is a function \( H_\phi \) that maps a noised MDF \( \mathbf{F}_t(Q) \), a set of query points \( Q \), and a diffusion timestep \( t \) to the parameters \( \theta \) of a  Molecular Neural Field (MNF) \( \mathbf{F}_\theta \). 

To encode the input MDF \( \mathbf{F}_t(Q) \), we begin by drawing a bounding box around the molecule and partitioning the enclosed volume into a uniform \( c \times c \times c \) 3D grid. A fixed number of query points are uniformly sampled within each cell. For each query point \( \mathbf{q}_j \), we first apply a learned positional encoding to its 3D coordinate and then concatenate the resulting embedding with the corresponding noised field vector \( \mathbf{F}_t(\mathbf{q}_j) \). These encoded query points form the input tokens to the transformer-based hypernetwork. 

The diffusion timestep \( t \) is embedded using sinusoidal Fourier features, as in \cite{mildenhall2021nerf}, and broadcast across all tokens. The transformer processes the sequence using self-attention layers to capture spatial dependencies across grid cells, followed by MLPs. The final token representations are pooled and passed through linear projections to produce the complete set of MLP weights \( \theta \) for the downstream field \( \mathbf{F}_\theta \).

\textbf{From scalar fields to vector fields.} Recall from \S\ref{sec:methods-training} that the hypernetwork takes a noised \textit{vector} field as input, but the loss is computed on an output \textit{vector} field. One challenge with this formulation is that it introduces a mismatch between the model’s input and output during generation: the input to the hypernetwork is a direction field, but the output is a scalar field. To resolve this, note that the direction field can be recovered from the distance field as follows:
\begin{equation}
    \nabla_{\mathbf{q}} \frac{1}{2} \left( f^{(k)}(\mathbf{q}) \right)^2 
    = \nabla_{\mathbf{q}} \frac{1}{2} \left\| \mathbf{a}_{\textrm{nearest}}^{(k)} - \mathbf{q} \right\|^2 
    = \mathbf{a}_{\textrm{nearest}}^{(k)} - \mathbf{q} 
    = \mathbf{F}^{(k)}(\mathbf{q})
\end{equation}
Here, \(\mathbf{a}_{\textrm{nearest}}^{(k)}\) denotes the nearest atom of type \(k\) to point \(\mathbf{q}\). This allows us to compute \(\mathbf{F}^{(k)}(\mathbf{q})\) from the predicted scalar field using automatic differentiation. During generation, we apply this transformation at each timestep to produce the direction field needed for the next step of the reverse diffusion. 

\textbf{Training curriculum.} We train with a curriculum on the noise level. For the first 100 epochs of training, the maximum noise level $t$ that we add is 10 (out of a total of $T = 1000$ noise levels). For every subsequent epoch, the maximum possible noise level is increased by 1, until $T = 1000$ is reached; we find that this accelerates training in practice; we show ablation studies in \S\ref{sec:ablation}

\textbf{Computational Complexity.} Since our method implicitly models the molecule via the MDF, the computational complexity scales only with the number of query points, not the size of the molecule. During training and inference, we keep the number of query points constant so memory complexity is constant.

\textbf{Hyperparameters.}  
We use the LAMB optimizer \cite{lamb} with a learning rate of \(1 \times 10^{-2}\), a batch size of 5500, and dropout set to 0.1. Both the noise schedule and the learning rate follow a cosine decay. The hypernetwork consists of 26 transformer layers and contains approximately 52 million parameters. All models are trained on NVIDIA H100 GPUs for 3500 epochs.

\begin{figure}
    \centering
    \caption{Overview of how we use \hdf{} as a foundational model for property prediction. In practice, we always extract the central activation within the stack of transformer layers, e.g., given 26 transformer blocks, we extract the 14th intermediate activation.}
    \includegraphics[width=\linewidth]{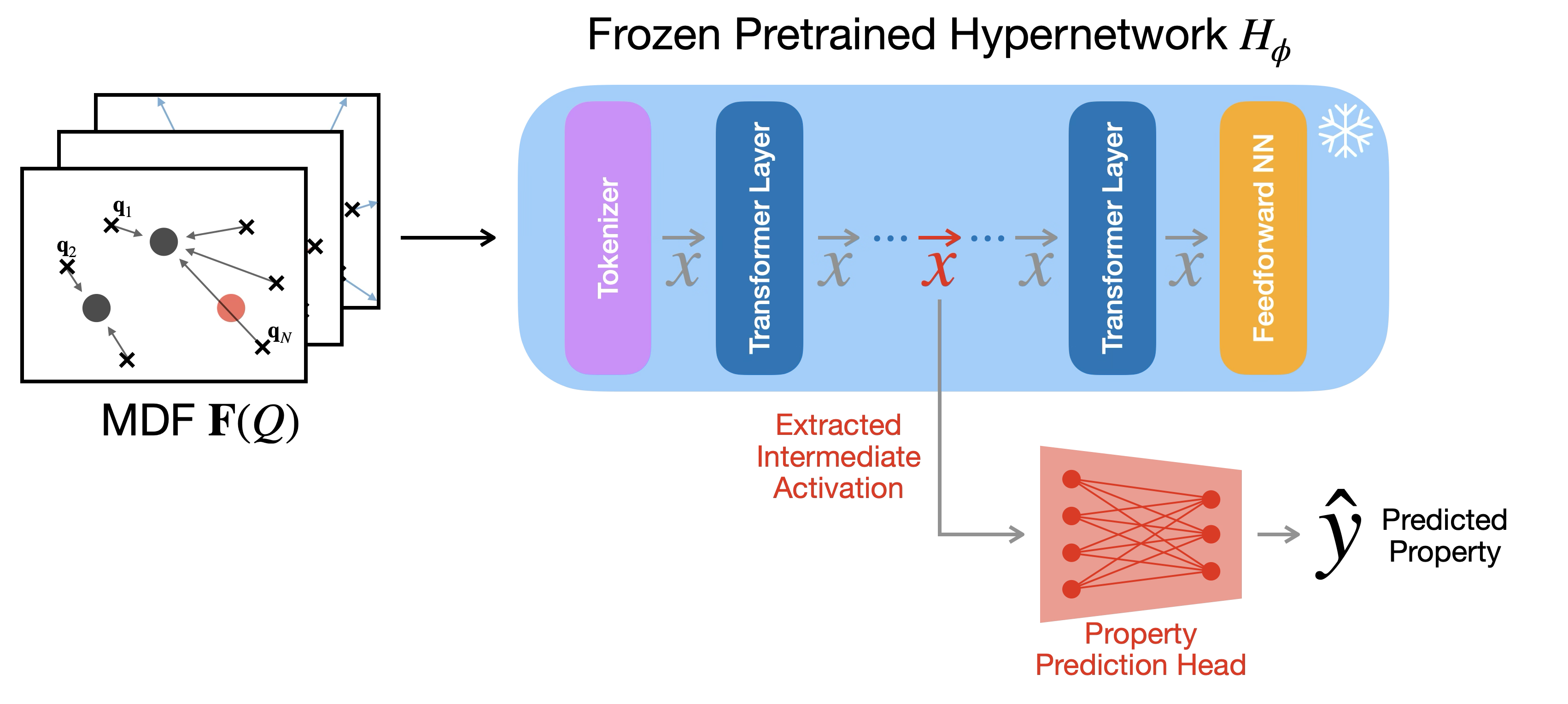}
    \label{fig:feature-extraction}
\end{figure}

\textbf{\hdf{} as a feature extractor} We illustrate how we leverage internal representations of \hdf{} as feature representations of molecules for downstream molecular property prediction in Figure \ref{fig:feature-extraction}. We pass an unnoised MDF for the molecule of interest into a frozen pretrained hypernetwork and extract an intermediate activation, which we average over the per-cell tokens. We then pass this activation into a property prediction head, which for us is a two layer MLP with $\tanh$ activations and hidden layer size 512. When training on MoleculeNet tasks, since MoleculeNet datasets represent molecules as SMILES strings rather than conformers, we generate a conformer for each SMILES string using ETKDG.

\section{Property Prediction Tasks}
In Table \ref{tab:moleculenet}, we provide a brief summary of the MoleculeNet tasks run in \S\ref{sec:results-property-prediction}.
\label{sec:property-prediction-tasks}
\begin{table}
    \centering
    \small
    \caption{Overview of the subset of MoleculeNet tasks that we perform property prediction on.}
    \begin{tabular}{lccl}
    \toprule
    \textbf{Task Name} & \textbf{Task Type}           & \textbf{Train Size} & \textbf{Description}                      \\ \midrule
    BBBP               & classification        & 1,631               & blood-brain barrier penetration          \\
    BACE               & classification        & 1,210               & human $\beta$-secretase 1 binding        \\
    Lipo               & regression            & 3,360               & lipophilicity                            \\
    ESOL               & regression            & 902                 & water solubility                         \\
    FreeSolv           & regression            & 513                 & hydration free energy                    \\
    \bottomrule
    \end{tabular}

    \label{tab:moleculenet}
\end{table}

\section{From MDFs to Molecules}
\label{sec:reconstruction}
While the output of \hdf{} is an MDF for a conformer, at inference time it is more desirable to have a SMILES or graph representation of a molecule. Our method of parsing MDFs into molecules involves two steps: (1) recovering atomic point clouds from MNFs and (2) recovering molecular bond information from atomic point clouds.

To recover atomic coordinates from MNFs, we find minima of the generated distance field and use these minima as atomic coordinates. We do this by sampling 4096 random query points in the volume of a molecule, computing the gradient of the predicted field at these query points, and using this gradient information to iteratively find the minima of field until a tolerance is met.

To recover molecular bond information from atomic point clouds, we use a combination of the standard chemoinformatics toolkits RDKit \cite{rdkit} and OpenBabel \cite{obabel} bond determination algorithms to convert atomic point clouds into RDKit-parsable molecules.

\section{Molecular Inpainting and Scaffold-Constrained Generation}
\label{sec:supp-masked-diff}
\begin{figure}
    \centering
    \includegraphics[width=0.75\textwidth]{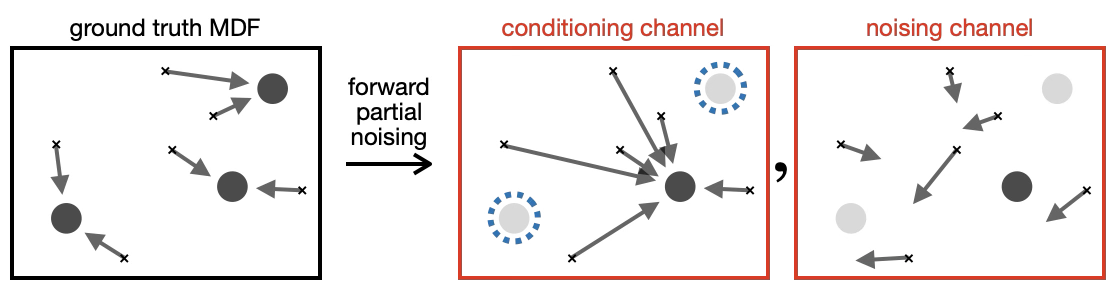}
    \caption{We extend our method to support conditional generation by expanding the input to two channels: the first encodes the field corresponding to the conditional atom, while the second initializes the diffusion process from pure noise.}
    \label{fig:partial-noising}
\end{figure}

Our training pipeline for partial noising is identical to that in \S\ref{sec:methods-training}, except in addition to a noised MDF, we also include a \textit{conditioning MDF} as an input to the hypernetwork. The conditioning MDF is computed by randomly deleting a subset of atoms from a conformer and recomputing the MDF for the partial molecule. This allows the network to learn to denoise a fully noised MDF \textit{conditioned on} a partial molecule being held constant (see Figure \ref{fig:partial-noising}).

\section{Scaling to Larger Biomolecules}
\label{sec:protein-scaling}
\hdf{} scales to large biomolecules in a way that current baseline architectures cannot. Most molecular generative models—such as EDM, GeoLDM, and VoxMol—are limited to small molecules with fewer than 100 heavy atoms. This is primarily due to architectural bottlenecks. GNN-based models like EDM and GeoLDM construct fully connected molecular graphs, resulting in \(\mathcal{O}(n^2)\) pairwise interactions for \(n\) atoms, which becomes computationally expensive as molecular size grows. VoxMol represents molecules using dense 3D voxel grids, which are memory-intensive and do not scale well to larger biomolecules. As shown in \cite{funcmol}, VoxMol fails to scale to the CREMP dataset \cite{cremp}, which includes macrocyclic peptides with an average of 74 heavy atoms.

\begin{figure}
    \centering
    \includegraphics[width=0.8\textwidth]{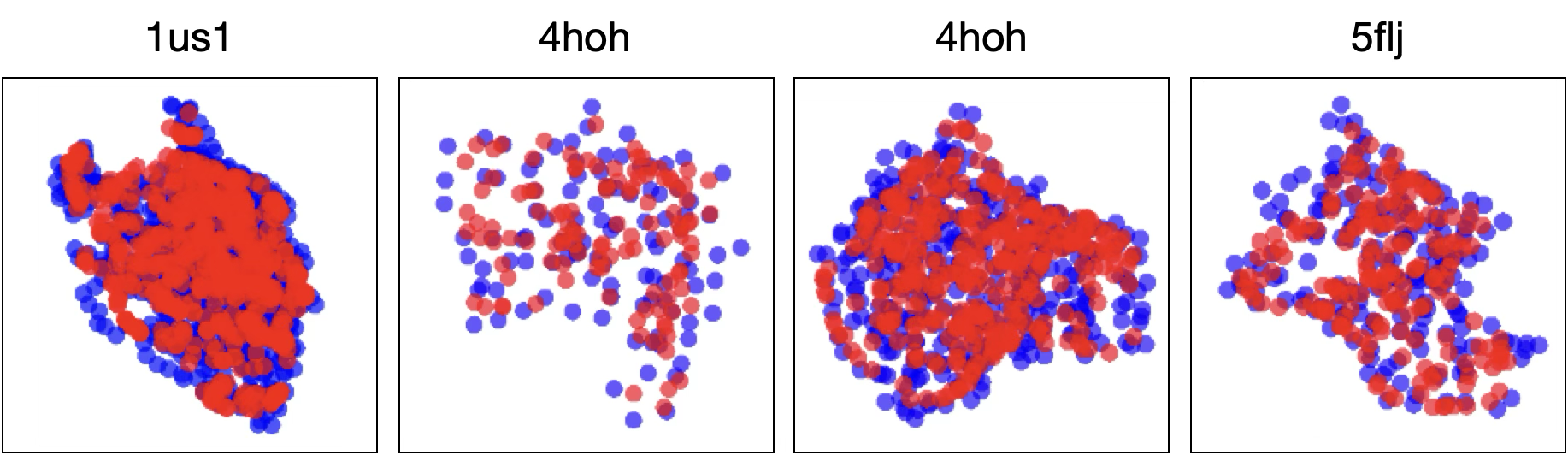}
    \caption{Four uncurated comparisons of ground truth alpha carbon coordinates of proteins from PLINDER (in blue) compared with  coordinates reconstructed from denoising a noised MDF (in red). We see that in all four cases, we are able to reconstruct the overall shape of the protein. This illustrates the scalability of \hdf{} to large biomolecules. Subplot captions are Protein Data Bank identifiers.}
    \label{fig:plinder-memorize}
\end{figure}

Proteins contain hundreds to thousands of atoms.  To evaluate the scalability of \hdf{}, we train our diffusion pipeline (see \S\ref{sec:methods-training}) on alpha carbon traces from the PLINDER dataset \cite{plinder}, a collection of 3D protein structures. Figure~\ref{fig:plinder-memorize} shows reconstructed proteins from the training set, generated by noising and denoising their molecular direction fields. In each case, we are able to recover the overall structure of the protein, demonstrating that \hdf{} can model large-scale molecular geometry.

The key insight enabling scalability is that the computational cost of \hdf{} grows quadratically with the number of query points, rather than the number of atoms in the molecule. This decouples computational complexity from molecular size: we train on large proteins by sampling a sparse set of query points, as the model sees each protein multiple times over training.

\section{Protein Feature Visualization}
\begin{figure}
    \centering
    \includegraphics[width=\textwidth]{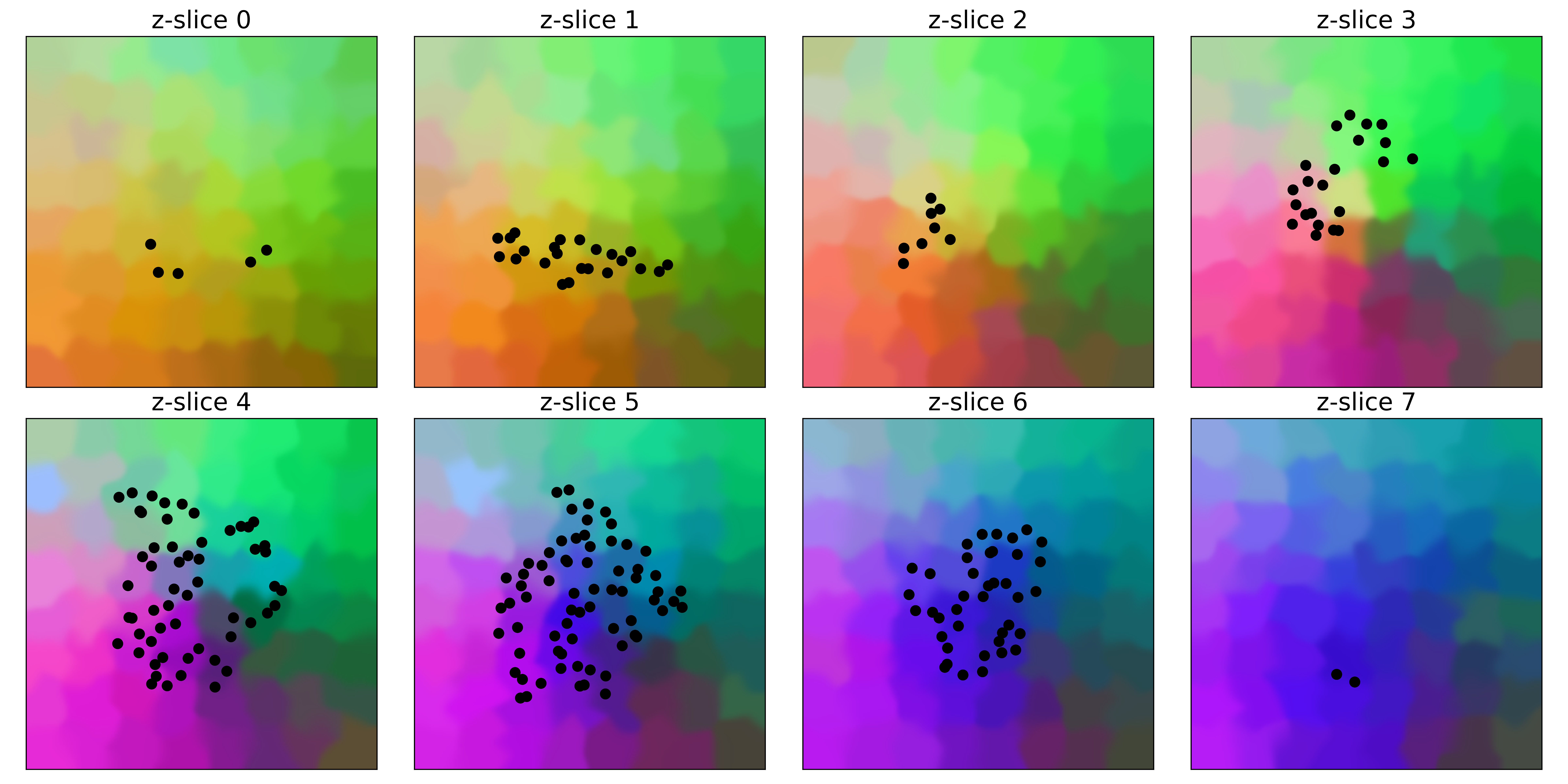}
    \caption{Visualization of spatially localized features for a protein (whose alpha carbons are plotted in black). The volume inhabited by the protein is separated into 8 $z$-slices. The latent feature corresponding to each query point is projected and scaled into $[0, 1]^3$ and interpreted as an RGB values; individual pixels of our visualization are colored by a weighted interpolation of neighboring query points.}
    \label{fig:protein-features}
\end{figure}

In Figure \ref{fig:protein-features}, we demonstrate how \hdf{} is able to produce spatially localized features for molecules. We input a protein structure through a pretrained \hdf{} and extract the intermediate activations from the hypernetwork. We are able to visualize the features at the per-query point resolution, after projecting from a high dimensional latent space to $\mathbb{R}^3$. The features smoothly change gradually in space.  

\section{Algorithms for Training and Generation}
\label{sec:algos}
Here we present pseudocode for training \hdf{} (Algorithm \ref{alg:training}) and generating sample conformers after training (Algorithm \ref{alg:generation}).

Algorithm \ref{alg:training} illustrates the noise scale curriculum we employ in training. For the first 100 epochs, we cap the maximum noise level $T$ at $10/1000$; for each epoch beyond the first 100, we increase the maximum noise level by 1 until we reach $T = 1000$.
\begin{algorithm}
\caption{Training Loop for \hdf{}}
\label{alg:training}
\begin{algorithmic}
\State $Q \gets$ random query points
\State $\mathbf{F}(Q) \gets$ ground truth MDF
\For{$\epoch = 1, 2, \dots$}
    \If{$\epoch \leq 100$}
        \State $T \gets 10$
    \Else
        \State $T \gets \max(\epoch - 100 + 10, 1000)$
    \EndIf
    \State $t \gets$ random uniform sample from $\{0, 1, \dots, T\}$
    \State $\alpha_t \gets$ cosine noise schedule
    \State $\boldsymbol{\eps} \gets\mathcal{N}(0, I)$\Comment{sample Gaussian noise}
    \State $\mathbf{F}_t(Q) \gets \sqrt{\overline{\alpha}_t} \cdot \mathbf{F}(Q) + \sqrt{1 - \overline{\alpha}_t} \cdot \boldsymbol{\eps}$\Comment{noise the MDF}
    \State $\theta \gets H_\phi(\mathbf{F}_t(Q), Q, t)$\Comment{pass noised MDF into $H_\phi$ to get predicted MNF parameters}
    \State $\mathcal{L}_{\textrm{dist}} \gets \frac{1}{|Q|} \sum_{\mathbf{q} \in Q} \sum_{k=1}^K \left| f^{(k)}_\theta(\mathbf{q}) - f^{(k)}(\mathbf{q}) \right|$\Comment{compute loss against ground truth distance field}
    \State $\phi \gets$ gradient update \Comment{update hypernetwork parameters from $\mathcal{L}_\textrm{dist}$}
\EndFor
\end{algorithmic}
\end{algorithm}

Algorithm \ref{alg:generation} illustrates how we generate novel molecules from a trained \hdf{}. We start by sample Gaussian noise and taking that to be a molecule noised at time step $t = 1000$. We then ask the hypernetwork to denoise it to $t = 0$, then noise the denoised prediction by noise level $t = 999$ and repeat. Since the outputs of the hypernetwork are the weights to an implicit \textit{distance field}, we must first compute the gradients with respect to the query points to convert it back to a \textit{direction field} that we can pass back into the hypernetwork.
\begin{algorithm}
\caption{Generation Procedure for \hdf{}}
\label{alg:generation}
\begin{algorithmic}
\State $t\gets 1000$
\State $Q\gets$ random query points
\State $F_t(Q)\gets$ Gaussian noise
\For{$t = 1000, 999,\dots, 1$}
    \State $\theta\gets H_\phi(F_t(Q), Q, t)$
    \State $\mathbf{F}_\theta(Q) \gets  \nabla_{Q} f_\theta(Q)$
    \State $\alpha_{t-1} \gets$ cosine noise schedule
    \State $\mathbf{F}_{t-1}(Q) \gets \sqrt{\overline{\alpha}_{t-1}} \cdot \mathbf{F}_\theta(Q) + \sqrt{1 - \overline{\alpha}_{t-1}} \cdot \boldsymbol{\eps}$
\EndFor
\end{algorithmic}
\end{algorithm}

\section{Ablation Studies}
\label{sec:ablation}

We conduct two ablation experiments to better understand key design choices in \hdf{}: (1) the impact of using a noise-level curriculum during training, and (2) the effect of the input field type provided to the hypernetwork.

\begin{figure}
    \centering
    \includegraphics[width=0.65\textwidth]{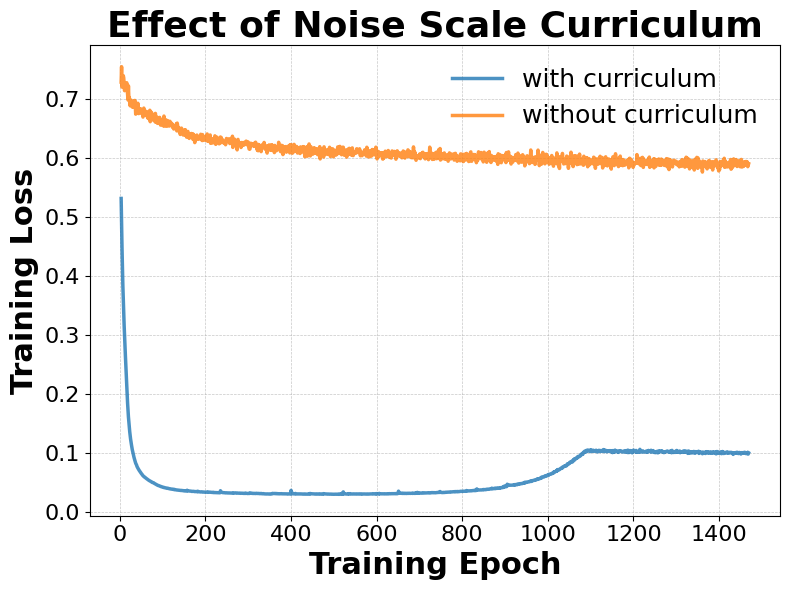}
    \caption{Ablation study illustrating the benefit of using a training curriculum (blue) on the diffusion noise scale versus no curriculum (orange). The curriculum improves convergence and early performance by exposing the model to gradually harder denoising tasks. The rise in loss near epoch 800 reflects the model being trained on highly noised inputs.}
    \label{fig:curriculum-ablation}
\end{figure}

\textbf{Effect of Curriculum on Noise Scale.} In Figure~\ref{fig:curriculum-ablation}, we compare training with and without a noise-level curriculum. In the curriculum setting, the maximum diffusion noise level is gradually increased over the course of training, starting from low noise levels and eventually reaching pure noise. This strategy leads to faster convergence and improved training stability compared to the baseline with a fixed maximum noise level throughout. Notably, we observe a rise in training loss around epoch 800 in the curriculum setting, which coincides with the introduction of highly noised inputs that are intrinsically more difficult to denoise.

\begin{figure}
    \centering
    \includegraphics[width=0.65\textwidth]{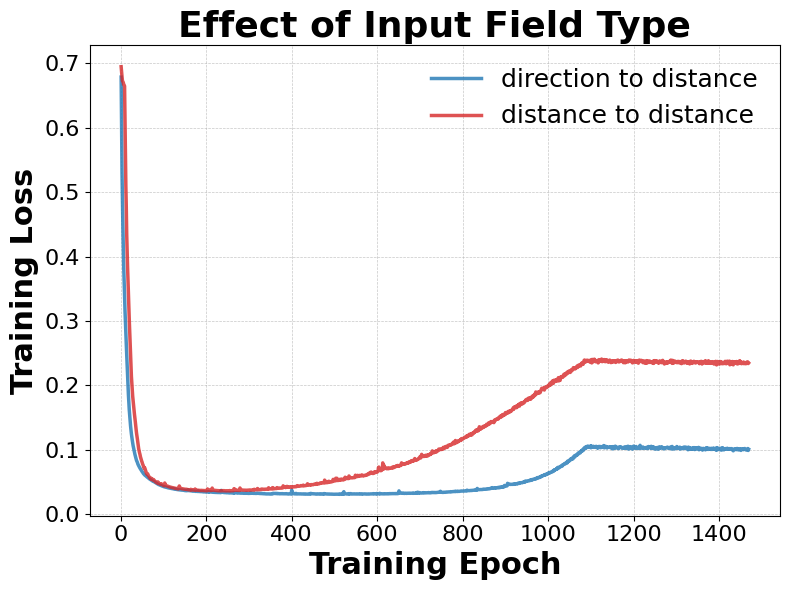}
    \caption{Ablation study comparing the effect of input field type. Using a \textit{direction} field (blue) as input leads to improved training over a \textit{distance} field (red), likely due to the richer high-frequency information present in direction fields.}
    \label{fig:field-type-ablation}
\end{figure}

\textbf{Effect of Input Field Type.} In Figure~\ref{fig:field-type-ablation}, we assess the impact of the type of input field—either a distance field or a direction field—provided to the hypernetwork. We find that using a direction field as input leads to lower training loss and more effective learning. We attribute this improvement to the fact that direction fields contain higher-frequency signals compared to distance fields, providing a more expressive representation of local molecular structure.

In both experiments, we report $\ell_1$ loss on the subset of query points that are \textit{very close} to the ground truth atomic coordinates—defined as the top 2\% of query points with the smallest distance to any atom. These points represent the highest-frequency components of the MDF, and accurately modeling them is critical for capturing precise atomic structure.

